\documentclass[11pt,a4paper]{article}
\usepackage{times,latexsym}
\usepackage{url}
\usepackage[T1]{fontenc}

\usepackage[acceptedWithA]{tacl2018v2}

\usepackage{microtype}
\usepackage{graphicx}
\usepackage{diagbox}

\usepackage{booktabs} 
\usepackage{caption}
\usepackage{subcaption}
\usepackage{multirow}
\usepackage{amssymb}  
\usepackage{amsmath}  
\usepackage{algorithm2e}  
\usepackage{amsthm}
\theoremstyle{definition}

\usepackage{hyperref}

\newcommand{\comment}[1]{}

\newcommand{ \our  }{\textsc{DISC}\xspace}

\usepackage{xspace,mfirstuc,tabulary}

\newif\iftaclinstructions
\taclinstructionsfalse 
\iftaclinstructions
\renewcommand{\confidential}{}
\renewcommand{\anonsubtext}{(No author info supplied here, for consistency with
TACL-submission anonymization requirements)}
\newcommand{\instr}
\fi

\iftaclpubformat 

\else

\fi

\title{Idiomatic Expression Identification using Semantic Compatibility}

 \author{Ziheng Zeng  \and Suma Bhat \\
 Department of Electrical and Computer Engineering\\
        University of Illinois at Urbana-Champaign\\
        Champaign, IL USA\\
        {\sf \{zzeng13, spbhat2\}@illinois.edu}}

\date{}

\begin{document}
\maketitle
\begin{abstract}
Idiomatic expressions are an integral part of natural language and constantly being added to a language. Owing to their non-compositionality and their ability to take on a figurative or literal meaning depending on the sentential context, they have  been a classical challenge for NLP systems. To address this challenge, we study the task of detecting whether a sentence has an idiomatic expression and localizing it when it occurs in a figurative sense. Prior art for this task had studied specific classes of idiomatic expressions offering limited views of their generalizability to new idioms. We propose a multi-stage neural architecture with attention flow  as a solution. The network effectively  fuses  contextual  and lexical information at different levels using word and sub-word representations. Empirical evaluations on three of the largest benchmark datasets with idiomatic expressions of  varied syntactic patterns and degrees of non-compositionality show that  our proposed model achieves new state-of-the-art results. A salient feature of the model is its ability to  identify idioms unseen during training with gains from $1.4\%$ to $30.8\%$  over competitive baselines on the largest dataset.

\end{abstract}

\section{Introduction} \label{sec: intro}

{Idiomatic expressions (IEs) are a special class of multi-word expressions (MWEs) that  typically occur as collocations and exhibit  \textit{semantic} non-compositionality (a.k.a. semantic idiomaticity), where the meaning of the expression is not derivable from its parts \cite{DBLP:reference/nlp/BaldwinK10}.} 
In terms of occurrence, IEs are individually rare, but collectively frequent in and  constantly  added to natural language across different genres \cite{moon1998fixed}. Additionally, they are known to enhance  fluency and used to convey ideas succinctly when used in everyday language  \cite{DBLP:reference/nlp/BaldwinK10,moon1998fixed}.

\begin{table}[t]
\centering
\caption{Example input and output for  \our  framework.  When a potentially idiomatic expression (PIE; italicized) is used {\color[HTML]{034f84}idiomatically}, the PIE is identified and extracted as the output; otherwise  the model outputs the start (\textless{}CLS\textgreater{}) and end  (\textless{}SEP\textgreater{}) tokens to indicate that it is used {\color[HTML]{d64161}literally}.}

\begin{tabular}{|p{12mm}|p{55mm}|}
\hline
\multicolumn{1}{|c|}{\multirow{2}{*}{\textbf{Input}}} & Tom said many bad things about Jane {\color[HTML]{034f84}\textit{behind her back}}. (Figurative)\\ \cline{2-2} 
\multicolumn{1}{|c|}{}                                            & He took one from an armchair and put it {\color[HTML]{d64161}\textit{behind her back}}. (Literal)          \\ \hline
\multirow{2}{*}{\textbf{Output}}                        & behind her back                                                     \\ \cline{2-2} 
                                                                  & \textless{}CLS\textgreater{} \textless{}SEP\textgreater{}                         \\ \hline
\end{tabular}

\label{tab: example_input_output}
\end{table}

 Classically regarded as  a ``pain in the neck'' to idiom-unaware NLP applications  \cite{sag2002multiword} these phrases  are challenging for reasons including their non-compositionality (semantic idiomaticity), besides taking a figurative or literal meaning depending on the context (semantic ambiguity), as shown by the example in Table~\ref{tab: example_input_output}.  
 Borrowing the terminology from \citet{haagsma2020magpie}, we call these phrases \textit{potentially idiomatic expressions (PIEs)} to account for the contextual semantic ambiguity. 
Indeed, prior work has identified the challenges PIEs pose to many NLP applications, such as machine translation \cite{fadaee2018examining, salton2014empirical}, 
  paraphrase generation \cite{ganitkevitch2013ppdb}, and sentiment analysis \cite{liu2017idiom, biddle2020leveraging}. 
Accordingly, making  applications idiom-aware, either by identifying them before or during the task, has been found to be effective \cite{korkontzelos2010can,nivre2004multiword,nasr-etal-2015-joint}.  
{This study proposes a novel architecture that detects the presence of a PIE.  When found,  its span in a given sentence is localized and returning  the phrase if it is  used figuratively (i.e., used as an IE); otherwise an empty string is returned indicating that the phrase is used literally (see Table~\ref{tab: example_input_output}).
Such a network can serve as a preprocessing step for broad-coverage downstream NLP applications because we consider the ability to detect IEs to be a  first step towards their accurate processing.
This is the \textit{idiomatic expression identification} problem, which  is  the MWE identification problem defined by \citet{DBLP:reference/nlp/BaldwinK10} limited to MWEs with semantic idiomaticity. 
}



Despite being well-studied in the current literature  as idiom type- and token classification  (e.g., \citet{fazly2009unsupervised,feldman2013automatic,salton2016idiom, taslimipoor2018identification, peng2018classifying, liu2019generalized}), previous methods are limited for various reasons.  They rely on knowing the PIEs  being classified  and hence {their} exact positions, or focus on specific syntactic patterns (e.g., verb-noun compounds or verbal MWEs), thereby calling into question their use in  more realistic scenarios with unseen PIEs (a likely event, given the prolific nature of PIEs). 
Additionally,   without a cross-type (type-aware) evaluation, where  the PIE types from the train and test splits are segregated \cite{DBLP:conf/starsem/FothergillB12,taslimipoor2018identification}, the true generalizability of these methods to unseen idioms cannot be inferred.
For instance, a model could be classifying by memorizing known PIEs or their tendencies to occur exclusively as figurative or literal expressions.


In contrast, this study aims to identify IEs in general (i.e., \textit{without} posing constraints on the PIE type) in a more realistic setting where  new idioms may occur,    by proposing the i\textbf{D}entifier of \textbf{I}diomatic expressions via \textbf{S}emantic \textbf{C}ompatibility (DISC) that performs detection and localization jointly. 
The novelty is that 
we perform the task without an explicit mention of the identity or the position of the PIE.  
As a result, the task  is   more challenging than the previously explored idiom token classification. 

An effective solution to this task calls for  the ability to relate the meaning of its component words with each other (e.g., \citet{baldwin2005deep,mccarthy2007detecting}) as well as with the context \cite{liu2019generalized}. This aligns with the widely upheld psycholinguistic findings on human processing of a phrase's figurative meaning in comparison with  its literal interpretation \cite{bobrow1973catching}. 
Toward this end, 
we rely on the contextualized representation of a PIE (accounting both for its internal and contextual properties), hypothesizing that a figurative expression's contextualized representation should be different from that of its literal counterpart. We refer to this as its \textit{semantic compatibility} (SC)---if a PIE is semantically compatible with its context, then it is literal; if not, it is figurative. The idea of SC also captures the distinction between literal word combinations and idioms, in terms of the semantics encoded by both 
{\cite{jaeger1999nature}}                        
and the {related property of \textit{selectional preference} \cite{DBLP:journals/ai/Wilks75}}---the tendency for a word to semantically select or constrain which other words may appear in its association \cite{katz1963structure} successfully used for processing metaphors \cite{shutova2013statistical} and word sense disambiguation \cite{stevenson2001interaction}. 
We capture SC by  effectively fusing information from the input tokens' contextualized and literal word representations to then localize the span of PIE used figuratively. Here we leverage the idea of attention flow  previously studied in a machine comprehension setting 
\cite{seo2016bidirectional}. 


Our main contributions in this work are:\\
\noindent A \textbf{novel IE identification model, DISC,} that uses attention flow  to fuse lexical semantic information at different levels and discern the SC of  a PIE. Taking only a sentence as input and using only word and POS representations, it  simultaneously performs detection and {localization} of the PIEs used figuratively. To the best of our knowledge, this is the first such study on this task.

\noindent \textbf{Realistic evaluation:}
We include two novel aspects in our evaluation methodology. First, we consider a new and stringent performance measure for subsequence identification; the identification is successful if and only if every word in the exact IE subsequence is identified. Second, we consider type-aware evaluation so as to highlight a     model's generalizability to unseen PIEs regardless of syntactic pattern. 

\noindent \textbf{Competitive performance:} 
Using  benchmark datasets with  a variety of PIEs, we show \our\footnote{The implementation of \our will be available at \url{https://github.com/zzeng13/DISC}} compares favorably  with strong baselines on PIEs seen during training. Particularly noteworthy is  its identification accuracy on \textit{unseen} PIEs, which  is 1.4\% to 11.1\% higher than the best baseline. 



\section{Related Work} \label{sec: related_work}
We provide a unified view of the diverse terminologies and tasks studied in prior works that define the scope of our study.

{\noindent \textbf{MWEs, IEs and metaphors.} 
We first introduce the relation between the three related concepts, namely MWE, IE, and metaphor, in order to present a clearer picture of the scope of our work.  
{According to \citet{DBLP:reference/nlp/BaldwinK10} and \citet{constant2017multiword},  MWEs (e.g., \textit{bus driver} and \textit{good morning}) satisfy the properties of outstanding collocation and contain multiple words. IEs are a special type of MWE that  also exhibit non-compositionality at the semantic level.}
{This has generally  been considered to be  the key distinguishing property between idioms (IEs) and  MWEs in general, although the boundary between IEs and non-idiom MWEs is not clearly defined. \cite{DBLP:reference/nlp/BaldwinK10, fadaee2018examining, liu2017idiom, biddle2020leveraging}.}
Metaphors are a form of figurative speech used to make an implicit comparison at an attribute level between two things  seemingly unrelated on the surface. By definition, certain MWEs and IEs use metaphorical figuration (e.g., \textit{couch potato} and \textit{behind the scenes}). However, not all metaphors are IEs because metaphors are not required to possess any of the properties of IEs, i.e., the components of a metaphor need \textit{not}  co-occur frequently (metaphors can be uniquely created by anyone), metaphors can be direct and plain comparisons and thus are not semantically non-compositional, and they need not have multiple words (e.g., \textit{titanium} in the sentence ``I am titanium").
}

\noindent \textbf{PIE and MWE processing.}
Current literature considers idiom type classification and idiom token classification \cite{cook2008vnc, liu2019generalized, liu2019toward} as two idiom-related tasks.  Idiom type classification decides if a phrase could be used as an idiom without specifically considering its context. Several works (e.g., \citet{fazly2006automatically, shutova2010metaphor}) 
have studied the  distinguishing properties of idioms from other literal phrases, especially that of  non-compositionality  \cite{westerstaahl2002compositionality, tabossi2008processing, tabossi2009idioms,reddy2011empirical,cordeiro2016predicting}. 

In contrast, idiom token classification \cite{fazly2009unsupervised,feldman2013automatic, DBLP:conf/simbig/PengF16a, salton2016idiom, taslimipoor2018identification, peng2018classifying, liu2019generalized} determines whether a given PIE is used  literally or figuratively in a  sentence. Prior works have used per-idiom classifiers {that are} completely non-scalable to be practical \cite{rajani2014using, liu2017representations}, {required} the position of the PIEs in the sentence (e.g., \citet{liu2019generalized}), 
and {focused} only on specific PIE patterns, such as  verb-noun compounds \cite{taslimipoor2018identification}. Overall, available works for this task only disambiguate a given  phrase. In contrast,   we do not assume any knowledge of the PIE being detected; given a sentence, we detect whether there is a PIE and disambiguate its use.

PIEs being special types of MWEs, our task is related to  \textit{MWE extraction} and \textit{MWE identification} \cite{DBLP:reference/nlp/BaldwinK10}. As with idioms, MWE extraction takes a text corpus as input and produces a list of  \textit{new} MWEs (e.g., \citet{fazly2009unsupervised, evert2001methods, pearce2001synonymy, schone2001knowledge}). 
MWE identification takes a text corpus as input and locates \textit{all} occurrences of MWEs in the text at the token level, differentiating between their figurative and literal use  \cite{baldwin2005deep, katz2006automatic, hashimoto2006japanese, blunsom2007structured, sporleder2009unsupervised, fazly2009unsupervised,savary2017parseme}; the identified MWEs may or may not be known beforehand. \citet{constant2017multiword} group main MWE-related tasks into \textit{MWE discovery} and \textit{MWE identification}: MWE discovery is identical to MWE extraction, while the MWE identification here, different from Baldwin and Kim's definition, identifies only \textit{known} MWEs. Our task is identical to \citet{DBLP:reference/nlp/BaldwinK10}'s MWE identification and {\citet{savary2017parseme}'s verbal MWE identification} while focusing only on PIEs and we aim to both detect the presence of PIEs and localize IE positions (boundaries), regardless of whether the PIEs were  previously seen or not. Besides, like idiom type classification and MWE extraction, our approach also works for identifying new idiomatic expressions. 

Approaches to MWE identification fall into two broad types. 
(1) A tree-based approach by first constructing a syntactic tree of the sentence and then traversing a selective set of candidate subsequences (at a node) to identify idioms \cite{liu2017idiom}.
{However, since the construction of a  syntactic tree is itself affected by the presence of idioms \cite{nasr-etal-2015-joint, green-etal-2013-parsing}, the nodes may not correspond to an entire  idiomatic expression, which in turn can affect even a perfect classifier's ability to identify idioms precisely.}
{(2) Framing the problem as a sequence labeling problem for token-level idiomatic/literal labeling, similar to prior works \cite{jang-etal-2015-metaphor, mao2019end, gong2020illinimet, kumar2020character, su2020deepmet} on metaphor detection that label each token as a metaphor or a non-metaphor} {and \citet{schneider-smith-2015-corpus}'s approach to MWE identification by tagging tokens from a MWE with the same supersense tag. This tagging approach provides a finer control over  subsequence extraction and  is unrestricted by factors that could impact a  tree-based approach, and does not require the traversal of all possible subsequences in search of the candidate phrases. 
Our approach is similar to this in spirit but focused on PIEs. In particular, \citet{schneider-smith-2015-corpus} aim to tag all MWEs while making no distinction for the non-compositional phrases, whereas our work aims to only identify IEs from sentences containing PIEs.}

\noindent \textbf{Semantic Compatibility.}
Exploiting SC for processing idioms  has been considered in  rather restricted settings, where the identity {of a PIE} (and hence its position)  is known.  
For instance, \citet{liu2019generalized} used  SC to classify a given phrase in its context as literal/idiomatic. 
A corpus of annotated phrases was used to train  a linear classification layer to discriminate between phrases' contextualized and literal embeddings. { \citet{DBLP:conf/simbig/PengF16a} directly check the compatibility between the word embeddings of a PIE with the embeddings of its context words to perform the literal/idiomatic classification. \citet{jang-etal-2015-metaphor} used SC and the global discourse context  to detect the figurative use of a small list of candidate metaphor words.}   \citet{gong2017geometry} treated the phrase's respective context  as vector spaces and modeled the distance of the phrase from the vector space as an index of SC.  We extend these prior efforts to identify both the presence and the position of an IE using only a sentence as input without knowing the PIE.



\section{Method} \label{sec: method}
\begin{figure*}[ht]
\centering
\includegraphics[width=1.\textwidth]{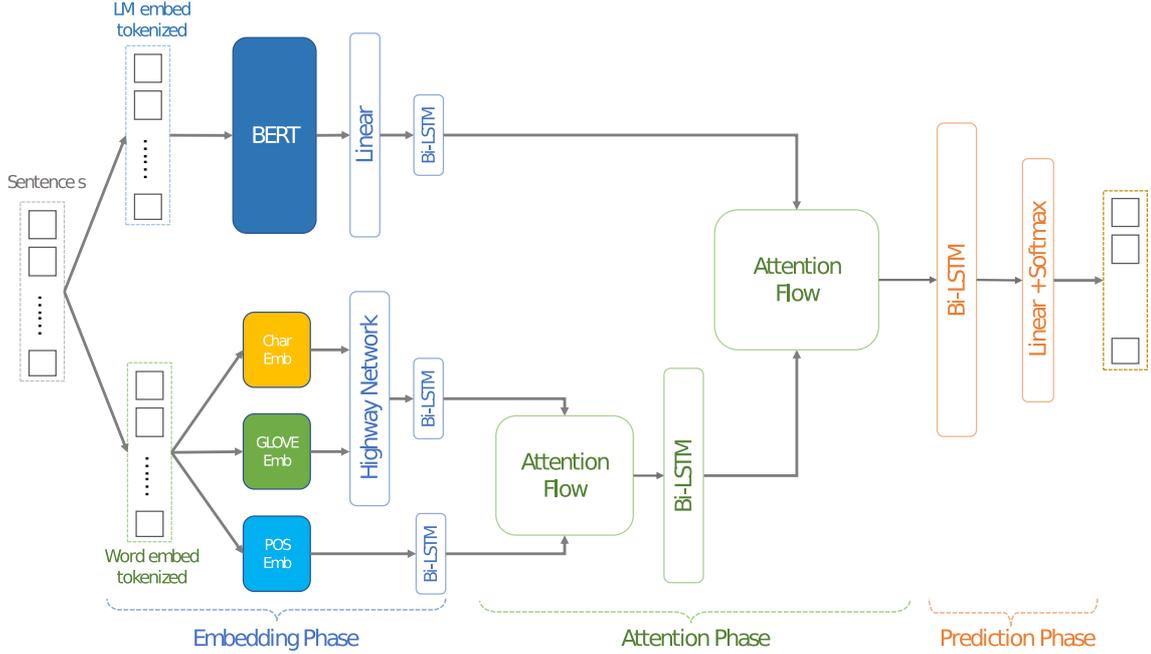}
\caption{Overview of the DISC framework.}
\label{fig: workflow}
\end{figure*}

In line with studies on MWE identification mentioned above, we frame the identification of idiomatic subsequences as a token-level tagging problem, where we perform literal/idiomatic classification for every token in the sentence. A simple post-processing step finally extracts the PIE subsequence used in the idiomatic sense. 

\noindent \textbf{Task Definition.}\label{subsec: overview}
Given an input  sentence $S = {w_1, w_2, ..., w_L}$, where $w_i$ for $i \in [1, L]$ are the {tokenized units} and  $L$ is the number of tokens in $S$, the task is to label the individual token $w_i$ with a label $c_i \in \{\sf idiomatic, literal\}$ so that the final output is a sequence of classifications $C = {c_1, c_2, ..., c_L}$. For a correct prediction, the phrase $w_{i:j}$ in $S$ is idiomatic and the corresponding $c_{i:j}$ are classified into the `idiom' class, while the rest are the `literal' class; or the phrase $w_{i:j}$ in $S$ is literal and the corresponding  $c_{1:L}$ are all classified into the `literal' class.


\noindent\textbf{Overview of proposed approach.} The overall workflow and model architecture of  \our   are illustrated in Figure \ref{fig: workflow}. The  model can be roughly divided into three distinct phases: (1) the embedding phase, (2) the attention phase, and (3) the prediction phase. In the embedding phase, the input sequence $S$ is tokenized and both the contextualized and static word embeddings are generated and supplemented 
with character-level information. Furthermore, POS tag embeddings  of the input tokens are generated to provide syntactic information. In the attention phase, an attention flow layer combines the POS tag embeddings with the static word embeddings yielding an enhanced literal representation for every word. Then, a second attention flow layer fuses the contextualized and the enriched literal representations by attending to the rich features of each token in the tokenized input sequence. Finally, the prediction phase further encodes the sequence of feature vectors and  performs token-level literal/idiomatic classification  to produce the predicted sequence $C$.


\noindent \textbf{Embedding Phase.}\label{subsec: data_preprocess}
Here the input sentence $S$ is  tokenized in two ways---one for the pre-trained language model and the other for the pre-trained static word embedding layer---resulting in two tokenized sequences $T^{c}$ and $T^{s}$, such that $|T^{c}| = M$ and $|T^{s}| = N$. Since the two tokenizers are not necessarily the same, $N$ and $M$ may be unequal. 

Next, $T^{c}$ is fed to a pre-trained language model to produce a sequence of contextualized word embeddings, $E^{con} \in \mathbb{R}^{M\times D_{con}}$, where $D_{con}$ is the embedding vector dimension.   A pre-trained word embedding layer takes $T^{s}$ to produce a sequence of static word embeddings, $E^{s} \in \mathbb{R}^{N\times D_s}$, where $D_s$ is the embedding vector dimension. {The contextualized embeddings capture the semantic content of the  phrases within the specific context, while the static word embeddings capture the compositional meaning of the phrases, both of which allow the model to check SC.}

Additionally, informed by the finding that character-level information alleviates the problem of morphological variability in idiom detection  \cite{liu2017idiom},  character sequences $C \in \mathbb{R}^{N\times W_t}$ are generated from  $T^{s}$, and their  character-level embeddings, $E^{char} \in \mathbb{R}^{N\times D_{char}}$ obtained using a 1-D Convolutional Neural Network (CNN) followed by a max-pooling layer over the maximum width of the tokens, $W_t$. 
 Then, $E^{char}$ and $E^{s}$ are combined via a two-layer highway network \cite{srivastava2015highway} which yields $\hat{E^{s}} \in\mathbb{R}^{N\times (D_{char}+D_{s})}$. 

Lastly, to capture shallow syntactic information, a POS embedding layer  generates a sequence of POS tags for $T^{s}$ and a simple linear embedding layer produces a sequence of POS tag embeddings, $E^{pos} \in \mathbb{R}^{N\times D_{pos}}$, where $D_{pos}$ is the POS embedding vector dimension. 

In effect,  the embedding layer encodes four levels of information: character-level, phrase-internal and implicit context (static word embedding), phrase-external and explicit context (contextual embedding), and shallow syntactic information (POS tag). 

To perform an initial feature extraction from the raw embeddings and unify the different embedding vector dimensions, we apply a Bidirectional LSTM (BiLSTM) layer for each embedding sequence  resulting in $\tilde{E^{con}} \in \mathbb{R}^{M\times D_{emb}} $, $\tilde{E^{s}} \in \mathbb{R}^{N\times D_{emb}} $, and $\tilde{E^{pos}} \in \mathbb{R}^{N\times D_{emb}} $, where $D_{embed}/2$ is the hidden dimension of the BiLSTM layers. 

\noindent \textbf{Attention Phase.}\label{subsec: data_preprocess}
The attention phase mainly consists of two attention flow layers. 
In its native application, i.e., reading comprehension, the attention flow layer linked and fused information from the context word sequence and the query word sequence \cite{seo2016bidirectional}, producing query-aware vector representations of the context words while propagating the word embeddings from the previous layer. Analogously, for our task, the attention flow layer fuses information from the two embedding sequences encoding  different kinds of information. More specifically, given two sequences $S^a \in \mathbb{R}^{L\times D}$ and $S^b\in \mathbb{R}^{K\times D}$ of lengths $L$ and $K$, the attention flow layer  computes $H\in \mathbb{R}^{L\times K}$ using, $H_{ij} = W_0^\top \left[S^a_{:i}; S^b_{:j}; S^a_{:i}\circ S^b_{:j} \right]$,
where $H_{ij}$ is the {attended,} merged embedding of the $i$-th token in $S^a$ and the $j$-th token in $S^b$, $W_0$ is a trainable weight matrix, $S^a_{:i}$ is the $i$-th column of $S^a$, $S^b_{:j}$ is the $j$-th column of $S^b$, $[; ]$ is vector
concatenation, and $\circ$ is the Hadamard product. Next, the attentions are computed from both $S^a$-to-$S^b$ and $S^b$-to-$S^a$. The $S^a$-to-$S^b$ attended representation is computed as  $\tilde{S^b}_{:i} = \sum_{j} a_{ij} S^b_{:j}$,
where $ a_i = \text{softmax} \left(H_{i:}\right)$; $a_i\in \mathbb{R}^K$ and $\sum a_{ij}=1$; $\tilde{S^b}\in \mathbb{R}^{2D\times L}$. The $S^b$-to-$S^a$ attended representation is computed as $\tilde{S^a}_{:i} = \sum_{i} b_{i} S^a_{:i}$,
where $b = \text{softmax} \left(\text{max}_{col}(H) \right)$, $b\in \mathbb{R}^L$, and $\tilde{S^a}\in \mathbb{R}^{2D\times L}$. Finally, the attention flow layer outputs a combined vector $U \in \mathbb{R}^{8D\times L}$, where $U_{:i} = [S^a_{:i};  \tilde{S}^b_{:i};  S^a_{:i}\circ  \tilde{S}^b_{:i};  S^a_{:i} \circ \tilde{S}^a_{:i}]$. 

The  \textit{two} attention flow layers serve different purposes. The \textit{first} one fuses the static word embeddings and the POS tag embeddings resulting in token representations that encode information from a given word's  POS  and that of its neighbors. The POS information is useful because different idioms often follow common syntactic structures (e.g., verb-noun idioms), which  can be used  to  recognize idioms unseen in the training data based on their similarity in syntactic structures (and thus aid generalizability). In all,  the first attention flow layer yields enriched static embeddings that more effectively capture the literal representation of the input sequence.
The \textit{second} attention flow layer combines the contextualized and literal embeddings so that the resulting representation  encodes the SC between the literal and contextualized representations of the PIEs. This is informed by  prior findings that the SC between the static and the contextualized representation of a phrase is a good indicator of its idiomatic usage \cite{liu2019generalized}. In addition, this attention flow layer permits working with  contextualized and static embedding sequences of differing lengths  using model-appropriate tokenizers for the pre-trained language model and the word embedding layer without having to explicitly map the tokens from the different tokenizers.

\noindent \textbf{Prediction Phase.}\label{subsec: data_preprocess}
The prediction phase consists of a single BiLSTM layer and a linear layer. The BiLSTM layer further processes and encodes the rich representations from the attention phase. The linear layer that follows uses a log softmax function to predict the probability of each token over the five target classes  {\sf idiomatic, literal, start, end}, and {\sf padding}. This architecture is inspired by the RNN-HG model from \cite{mao2019end} with the difference  that our BiLSTM  has only one layer. During training, the token-level  negative log-likelihood loss is computed and backpropagated to update the model parameters.    

\noindent \textbf{Implementation details.}\label{subsec: implementation}
 In our implementation, the tokenizer for the language model uses the WordPiece algorithm \cite{schuster2012japanese}   prominently used in BERT \cite{devlin-etal-2019-bert}, whereas  the static word embedding layer used Python's Natural Language Toolkit (NLTK) \cite{Loper02nltk}. 
 
The pre-trained language model is the uncased base BERT from Huggingface's Transformers package \cite{wolf-etal-2020-transformers} with an embedding dimension of $D_{con} = 768$. The pre-trained word embedding layer is the cased Common Crawl version of GloVE, which has a vocabulary of 2.2 M words and the embedding vectors are of dimension $D_{s} = 300$ \cite{pennington2014glove}. Both the BERT and GloVE models are frozen during training. We use  NLTK's POS tagger for the POS tags. 

For the character embedding layer, the input embedding dimension is 64 and the number of CNN output channels is $D_{char} = 64$. The highway network has two layers. The POS tag embedding is of dimension $D_{pos} = 64$. All the BiLSTM layers have a hidden dimension of 256, and thus $D_{emb} = 512$.

\section{Experiments} \label{sec: experiments}

\begin{table*}[t]
\centering
\caption{Statistics of the datasets in our experiments showing the size of training and testing sets, proportion of instances having a figurative PIE (pct. idiomatic), the size of the PIE set (\# of idioms), the average number of occurrences per PIE (avg. idiom occ), and standard deviation of the number of occurrences per PIE (std. idiom occ). }
\label{tab: datasets}
\resizebox{\textwidth}{!}{
\begin{tabular}{|l|l|l|l|l|l|l|l|l|l|}
\hline
\multirow{2}{*}{\textbf{Dataset}}   & \multirow{2}{*}{\textbf{Split}} & \multicolumn{2}{l|}{\textbf{Size (pct. idiomatic)}} & \multicolumn{2}{l|}{\textbf{\# of idioms}} & \multicolumn{2}{l|}{\textbf{Avg. idiom occ}} & \multicolumn{2}{l|}{\textbf{Std. idiom occ}} \\ \cline{3-10} 
                                    &                                 & \textbf{Train}            & \textbf{Test}           & \textbf{Train}       & \textbf{Test}      & \textbf{Train}        & \textbf{Test}        & \textbf{Train}        & \textbf{Test}        \\ \hline
\multirow{2}{*}{\textbf{MAGPIE}}    & \textbf{Random}                 & 32,162 (76.63\%)           & 4030 (76.48\%)           & 1,675                & 1,072              & 19.2                  & 3.76                 & 24.82                 & 3.65                 \\ \cline{2-10} 
                                    & \textbf{Type-aware}             & 32,155 (77.90\%)           & 4,050 (70.54\%)          & 1,411                & 168                & 22.79                 & 24.11                & 29.96                 & 32.05                \\ \hline
\multirow{2}{*}{\textbf{SemEval5B}} & \textbf{Random}                 & 1,420 (50.56\%)            & 357 (50.70\%)            & 10                   & 10                 & 142                   & 35.7                 & 51.25                 & 12.69                \\ \cline{2-10} 
                                    & \textbf{Type-aware}             & 1,111 (58.74\%)            & 341 (58.65\%)            & 31                   & 9                  & 35.81                 & 37.89                & 28.84                 & 30.12                \\ \hline
\multirow{2}{*}{\textbf{VNC}}       & \textbf{Random}                 & 2,285 (79.52\%)            & 254 (70.47\%)            & 53                   & 50                 & 43.11                 & 5.08                 & 25.89                 & 2.93                 \\ \cline{2-10} 
                                    & \textbf{Type-aware}             & 2,191 (79.69\%)            & 348 (71.84\%)            & 47                   & 6                  & 46.62                 & 58                   & 27.99                 & 27.77                \\ \hline
\end{tabular}
}
\end{table*}

\noindent \textbf{Datasets.} \label{subsec: datasets}
We use the following three of the largest available datasets of idiomatic expressions to evaluate the proposed model alongside other baselines. 
\noindent\textit{MAGPIE} \cite{haagsma2020magpie}: MAGPIE is a recent, the largest-to-date  corpus of PIEs in English. It consists of   1,756 PIEs across different syntactic patterns 
along with the  sentences in which they occur (56,622 annotated data instances with an average of 32.24 instances per PIE), where the sentences are drawn  from a diverse set of genres, such as news and science,  collected from resources such as the British National Corpus (BNC) \cite{BNC_CON}. 
For our experiments, we only considered the complete sentences of up to 50 words in length that contain the unambiguously labelled PIEs (as indicated by the perfect confidence score). 
\\
\noindent\textit{SemEval5B} \cite{korkontzelos2013semeval}: 
This set has 60 PIEs unrestricted by syntactic pattern appearing in 4,350 sentences from the ukWaC corpus \cite{baroni2009wacky}. As in MAGPIE, we only consider the sentences with the annotated phrases.  \\
\noindent\textit{VNC} \cite{cook2008vnc}: {Verb Noun Combinations (VNC) dataset is a popular benchmark dataset that contains expert-curated 53 PIE types that are only verb-noun combinations and around 2,500 sentences containing them either in a figurative or literal sense---all extracted from the BNC. Because VNC does not mark  the location of the idiom,  we manually labeled them. }\\
Together, the datasets account for a wide variety of PIEs, making this  the largest available study on a wide variety of PIE categories.



\noindent \textbf{Baseline models.} \label{subsec: baseline_models} {We use the following six baseline models for our experiments. We note that because our method is similar to idiom type classification only in its end goal and not in setting, we exclude SOTA models for idiom classification from this comparison, but include the more recent MWE extraction methods. }

\medskip
\noindent\textit{Gazetteer} is a na\"{i}ve baseline that looks up a PIE in a  lexicon. In our experiments, to make the Gazetteer method independent of the algorithm and lexicon, we present the theoretical performance upper bound for any Gazetteer-based algorithm as follows.  We assume that the Gazetteer perfectly detects the idiom boundaries in sentences and, in turn, predicts all PIEs to be  idiomatic. We point out that since the idiomatic class is the most frequent-class in all of our benchmark datasets, this also turns out to be the \textit{majority-class} baseline for the case of sentence-level, binary idiomatic and literal classification, {i.e., it predicts every sentence in a dataset to be idiomatic for binary idiom detection.}

\medskip
\noindent\textit{BERT-LSTM} has a simple architecture that combines the pre-trained BERT and a linear layer to perform a binary classification at each token and was used in  \cite{kurfali2020disambiguation}  for disambiguating PIEs. 

\medskip
\noindent\textit{Seq2Seq} has an encoder-decoder structure and is commonly used in sequence tagging  tasks \cite{filippova2015sentence, malmi-etal-2019-encode, dong-etal-2019-editnts}. It first uses the pre-trained BERT  to generate contextualized embeddings and then sends them to a BiLSTM encoder-decoder model to tag each token as literal/idiomatic. Although not commonly used in idiom processing tasks, the encoder-decoder framework serves as a simple yet effective baseline for our tagging based idiom identification.  

\medskip
\noindent\textit{BERT-BiLSTM-CRF} \cite{huang2015bidirectional} is an established model for sequence tagging (and the state-of-the-art  for name entity recognition in different languages \cite{huang2015bidirectional, hu2020named}), which uses a BiLSTM to encode the sequence information and then performs sequence tagging with a conditional random field (CRF). 

\medskip
\noindent\textit{RNN-MHCA} \cite{mao2019end} is a recent state-of-the-art model for metaphor detection on the  benchmark VUA dataset that uses GloVe and ELMo embeddings with a multi-head contextual attention. 

\medskip
\noindent\textit{IlliniMET} \cite{gong2020illinimet} is one of the most recent models for metaphor detection, achieving state-of-the-art performance on VUA \cite{VUA} and TOEFL \cite{beigman-klebanov-etal-2018-corpus} dataset. It uses RoBERTa and a set of linguistic features to perform token level metaphor tagging. 

\medskip
\noindent \textbf{Experimental setup.}
For a fair comparison across the models,  we use a pre-trained BERT model in place of the {linear} embedding layers, ELMo, and RoBERTa model respectively in the last three baselines. The pre-trained BERT model is also frozen for all the baseline model and \our. {Owing to a lack of a good fine-tuning strategy that fits all baselines, we leave to future work exploring improved performance via end-to-end BERT fine-tuning.}

{ In order to test the models' ability to identify unseen idioms, each dataset was split into train and test set in two ways: \textit{random} and \textit{type-aware}. In the random split, the sentences are randomly divided and the same PIE can appear in both sets, whereas in the type-aware split, the idioms in the test set and the train set do not overlap. For MAGPIE and SemEval5B, we use their respective random/type-aware and train/test splits. For VNC, to create the type-aware split,  we randomly split the idiom types by a 90/10 ratio, leaving 47 idiom types in train set and 6 idiom types in test set. }
For every dataset split, we trained every model for $600$ epochs with a batch size of $64$, an initial learning rate of $1e-4$, using the Adam optimizer.

The checkpoints with the best test set performance during training are recorded later in the result tables. For models with BiLSTMs, we used the same specifications as in our model with a hidden dimension of $256$ and a single layer, except for BiLSTM-CRF, where we used a stacked two-layer BiLSTM. 
For the linear layers, we set a dropout rate of $0.2$ during training. For Seq2Seq, we used a teacher forcing ratio of $0.7$ during training and brute force search during inference. The same pre-trained BERT model from Huggingface’s Transformers package was used as a frozen embedding layer in all models. All the other hyperparameters were in their default values. 

All training and testing were done on a single machine with an Intel\textsuperscript{\tiny\textregistered} Core\textsuperscript{\tiny\texttrademark} i9-9900K processor and a single NVIDIA\textsuperscript{\tiny\textregistered} GeForce\textsuperscript{\tiny\textregistered} RTX 2080 Ti graphic card. 

\noindent \textbf{Evaluation metrics.} \label{subsec: metrics} 
We use two metrics to evaluate the performance of the models.  (1) \textit{Classification F1 score}  (\textbf{F1}) measures the binary idiom detection performance at the sequence level {with the presence of idioms being the positive class}. (2) \textit{Sequence accuracy} (\textbf{SA}) computes the idiom identification performance at the sentence level, where a sequence is considered as being classified correctly if and only if all its tokens are tagged correctly. 
We point out that the performance in terms of F1 score is essentially analogous to the performance of the idiom token classification task (see Section~\ref{sec: related_work}), the primary difference being whether the idiom is specified or not.
Because SA is stricter than F1, we regard it to be the most relevant metric for idiom detection and span localization. Here we consider SA to be the primary evaluation metric with F1 providing additional performance references. 

\begin{table*}[t]
\centering
\small
\caption{Performance of models on the MAGPIE, SemEval5B, and VNC Dataset as evaluated by Classification F1 score (F1;\%) and Sequence Accuracy  (SA;\%); best performances are boldfaced; performances marked with asterisks are comparable in their differences are not statistically significant at $p=0.05$ using bootstrapped samples that are estimated $10^5$ times.}
\begin{tabular}{|l|l|r|r|r|r|r|r|}
\hline
\multirow{2}{*}{\textbf{Data Split}} & \multirow{2}{*}{\textbf{Model}} & \multicolumn{2}{c|}{\textbf{Magpie}}   & \multicolumn{2}{c|}{\textbf{SemEval5B}}   & \multicolumn{2}{c|}{\textbf{VNC}}                                   \\ \cline{3-8} 
                                     & & \multicolumn{1}{l|}{\textbf{F1}} & \multicolumn{1}{l|}{\textbf{SA}} & \multicolumn{1}{l|}{\textbf{F1}} & \multicolumn{1}{l|}{\textbf{SA}} & \multicolumn{1}{l|}{\textbf{F1}} & \multicolumn{1}{l|}{\textbf{SA}} \\ \hline\hline
\multirow{6}{*}{\textbf{Random}}     & Gazetteer                      & 86.67              & 76.47                  & 67.29                 & 50.70                         & 82.68                     & 70.47             \\ \cline{2-8}
                                     & BERT                           & 87.16              & 37.10                  & 92.51                 & 76.47                         & 93.09                     & 50.00             \\ \cline{2-8} 
                                     & Seq2Seq                        & 92.70              & 83.21                  & 94.41                 & $^*$94.12                     & 95.21                     & 86.61             \\ \cline{2-8} 
                                     & BERT-BiLSTM-CRF                & 94.22              & $^*$\textbf{87.71}     & 93.29                 & 92.44                         & 95.45                     & 85.03             \\ \cline{2-8} 
                                     & RNN-MHCA                       & \textbf{95.51}     & $^*$ 86.82             & $^*$ 94.94            & 93.56                         & $^*$96.15                 & 91.33             \\ \cline{2-8} 
                                     & IlliniMET                      & 86.54              & 37.97                  & 92.59                 & 78.15                         & 93.55                     & 59.45             \\ \cline{2-8} 
                                     &  \our                          & 95.02              & $^*$87.47              & $^*$\textbf{95.80}    & $^*$\textbf{95.23}            & $^*$\textbf{96.97}        & \textbf{93.31}    \\ \hline \hline
\multirow{6}{*}{\textbf{Type-aware}} & Gazetteer                      & 82.73              & 0.00                   & \textbf{73.94}        & 0.00                          & 83.61                     & 0.00              \\ \cline{2-8}
                                     & BERT                           & 86.27              & 39.70                  & 73.37                 & 35.19                         & 86.85                     & 50.86             \\ \cline{2-8} 
                                     & Seq2Seq                        & 83.81              & 63.42                  & 50.35                 & 44.28                         & 88.80                     & 73.56             \\ \cline{2-8} 
                                     & BERT-BiLSTM-CRF                & 80.47              & 61.78                  & 57.82                 & 44.57                         & 83.30                     & 65.52             \\ \cline{2-8} 
                                     & RNN-MHCA                       & 86.34              & 61.42                  & 56.25                 & 42.23                         & $^*$88.74                 & 79.02             \\ \cline{2-8} 
                                     & IlliniMET                      & 83.58              & 39.68                  & 69.49                 & 41.94                         & 87.97                     & 54.60             \\ \cline{2-8} 
                                     &  \our                          & \textbf{87.78}     & \textbf{70.47}         & 58.82                 & \textbf{55.71}                & $^*$\textbf{89.02}        & \textbf{80.46}    \\ \hline
\end{tabular}

\label{tab: perf_all_datasets}
\end{table*}

\section{Results and Analyses}
\noindent \textbf{IE identification performance.} \label{subsec: performance}
A comparative evaluation of the models on the MAGPIE, SemEval5B, and VNC datasets is shown in Table~\ref{tab: perf_all_datasets}.

Overall,  \our is the best performing model among all baseline models. Specifically, \our and RNN-MHCA show competitive results in all random split settings, however, \our  has stronger performance on type-aware settings, indicating 
that the SC check enables \our  to recognize the non-compositionality of  idioms permitting it to  generalize better to  idioms unseen in the training set. Therefore, while RNN-MHCA might be as good as \our  when it comes to identifying (and potentially memorizing) known idioms, \our is more capable of identifying unseen idioms since it better leverages the SC property of idioms in addition to memorization. 

In the random setting, \our performs on par with RNN-MHCA and BERT-BiLSTM-CRF in terms of F1 and SA for MAGPIE {while outperforming all baselines using the other datasets. It is notable that even with the ability to perfectly localize PIEs, Gazetteer has a low SA compared to the other top-performing models due to its inability to use the context to determine if the PIE is used idiomatically.} 
In the type-aware setting,  the F1 of \our is comparable to that of  RNN-MHCA and BERT-BiLSTM. However, in terms of SA, \our outperforms all models across all datasets.  
We also observe that for all datasets, achieving high  F1 scores is much easier than achieving high SA. This is especially salient in the MAGPIE type-aware split where all the models achieve similar F1s, whereas  {\our outperforms the others in terms of SA by margins ranging from $7\%$ to $30.8\%$ absolute points.} {Moreover,  Gazetteer is unable to perform PIE localization at all in this setting on accounts of its being limited to the instances available in an idiom lexicon. }

For MAGPIE random split, it is notable that all the models (including Gazetteer with its majority-class prediction) achieve at least $86\%$  F1 score. For MAGPIE type-aware split, \our is decisively the best performing model with absolute gains of at least $7.1\%$ in SA and at least $1.4\%$ in  F1. 
For SemEval5B type-aware split, \our  is the best performing model in terms of SA with gains of at least $11.1\%$. Note that in terms of F1, although BERT  outperforms \our  by $14.6\%$,  Gazetteer outperforms all methods. {We believe this is due to a combination of factors, including the insufficiency of the training instances (there were only 1,111 instances) and the number of idioms (there were only 31 unique idioms in the train set)},  and the distributional dissimilarity between the train and test sets  with respect to the semantic and the syntactic properties of the PIEs in the SemEval5B dataset, e.g., unlike VNC where both the train and test idioms were  verb-noun constructions, SemEval5B idioms are of more diverse syntactic structures, yet SemEval5B has fewer   training instances and total  number of idioms.  However, \our  outperforms BERT  by $20.5\%$ in SA, which shows that \our  has the best idiom identification ability. 
For VNC, \our  and RNN-MHCA perform competitively in all evaluation metrics in both random- and type-aware settings. In terms of SA, \our  has a $>1\%$ gain over RNN-MHCA in both random and type-aware settings.

\noindent \textbf{Idiom identification cross-domain performance across datasets.} \label{subsec: performance_cross-domain}
To check the cross-domain performance of the best performing models (\our  and RNN-MHCA), we train them  on the MAGPIE  train set (as it contains the largest number of instances) and test  on the VNC  and the SemEval5B  test sets in a random-split setting.  {As shown in Table \ref{tab: performance_cross-domain}, both models show a performance drop due to the change of the sentence source between SemEval5B and MAGPIE, and the small overlap between VNC and MAGPIE (only 4 common idioms)}. 
RNN-MHCA  obtains F1 scores that are 3.6\% and 1.44\% higher than that of \our on SemEval5B and VNC respectively, indicating its better ability to {detect} PIEs in this cross-domain setting. However, \our  is able to {detect} and locate the idioms more precisely, yielding SA gains of  7.8\% and 2.81\% over those of RNN-MHCA on SemEval5B and VNC respectively. We argue that this demonstrates  \our's  ability to identify idioms with a higher precision and that \our's gain in SA outweighs its loss in F1, given that the gain is generally higher than the loss and  SA is a more reliable  measure of identification performance. 


\begin{table}[]
\centering
\caption{The performance of idiom identification in a cross-domain setting where models are trained on MAGPIE random and tested on target domains (Tgt. Domain) SemEval5B random and VNC random. The performance is measured by  Classification  F1  score (F1;\%) and Sequence Accuracy (SA;\%).}
\resizebox{\columnwidth}{!}{
\begin{tabular}{|l|l|r|r|}
\hline
\textbf{Tgt. Domain}              & \textbf{Models} & \multicolumn{1}{l|}{\textbf{F1}} & \multicolumn{1}{l|}{\textbf{SA}} \\ \hline
\multirow{2}{*}{\textbf{SemEval5B}} & RNN-MHCA        & 81.35                           & 54.72                           \\ \cline{2-4} 
                                    & \our              & 77.70                           & 61.80                            \\ \hline
\multirow{2}{*}{\textbf{VNC}}       & RNN-MHCA        & 85.01                           & 69.74                           \\ \cline{2-4} 
                                    & \our              & 83.57                           & 72.55                           \\ \hline
\end{tabular}
}

\label{tab: performance_cross-domain}
\end{table}


We now evaluate the performance by paying specific attention to one specific property of PIEs that makes them challenging to NLP applications--- syntactic flexibility (fixedness) \cite{constant2017multiword}. 

\noindent \textbf{Effect of idiom fixedness.} \label{subsec: idiom_fixedness_level}
We analyze the idiom identification performance with respect to the idiom fixedness levels. Following the definitions given by Sag et al. \cite{sag2002multiword} for lexicalized phrases, we categorized idioms into three fixedness levels:  (1) \textit{fixed} (e.g. \textit{with respect to})---fully lexicalized with no morphosyntactic variation or internal modification, (2) \textit{semi-fixed}, (e.g. \textit{keep up
with})---permit limited lexical variations (\textit{kept up with})  such as inflection and determiner selection, while adhering to strict constraints on word order and composition, and (3) \textit{syntactically-flexible} {(e.g. \textit{serve someone right})}---largely retain basic word order and permit a wide range of syntactic variability such that the internal words of the idioms are subject to change. The authors (both near-native English speakers, one with linguistics background) manually labeled the PIEs in the MAGPIE test set  into these 3 levels by first independently labeling $35$ per level. Seeing that the agreement was $91\%$   all the remaining idioms were labeled by one researcher.  We note that the highest level is  occupied by verbal MWEs (VMWEs) that are characterized by  complex structures, discontinuities, variability, and ambiguity \cite{savary2017parseme}.

We use this labeled set to  compute the \our  performance for each fixedness level in terms of classification F1 and SA. As shown in Table \ref{tab: perf_vs_idiom_level}, although the fixed idioms obtain the best performance as expected, the performance difference between semi-fixed and syntactically flexible idioms suggests that \our can reliably detect idioms from different fixedness levels.

\begin{table}[]
\centering
\caption{Performance of \our  on MAGPIE random split dataset as evaluated by Classification F1 score (F1;\%) and Sequence Accuracy (SA;\%) for each idiom fixedness level.}
\label{tab: perf_vs_idiom_level}\
\resizebox{\columnwidth}{!}{

\begin{tabular}{|c|c|r|r|}
\hline
\multirow{2}{*}{\textbf{Metric}} & \multicolumn{3}{c|}{\textbf{Idiom fixedness level}}                                                                            \\ \cline{2-4} 
                                 & \textbf{Fixed}              & \multicolumn{1}{c|}{\textbf{Semi-fixed}} & \multicolumn{1}{c|}{\textbf{Syntactically-flexible}} \\ \hline
\textbf{SA}                 & \multicolumn{1}{r|}{93.11} & 88.25                                   & 87.57                                               \\ \hline
\textbf{F1}                  & \multicolumn{1}{r|}{94.75} & 92.10                                   & 92.66                                               \\ \hline
\end{tabular}
}
\end{table}


\noindent \textbf{Error analysis.}\label{subsec: performance_error_analysis}
Next, we analyze \our's performance and its errors on the MAGPIE dataset to gain further insights into \our's idiom identification abilities and its shortcomings. 

A closer inspection of the results showed that  65.9\% of the $1,071$ idiom types from the MAGPIE random split test set {have perfect average SA (i.e., 100\%)}, indicating that \our  successfully learned to recognize the SC of the vast majority of the idiom types from the training set. 

In order to gain insights  related to  \our's  ability to memorize the instances of known PIEs to perform identification on the known ones, we analyze the relationship between the average SA  and the number of training samples on a per PIE basis in the MAGPIE random split using Pearson correlation. A correlation of  $0.1857$ with a $p<0.05$ 
indicates a  weak relationship  between the number of training instances and the performance. This, taken together with the strong type-aware performance, suggests that \our's  identification ability relies on   more than just memorizing known PIE instances. 

Visualizing the attention matrices (matrix $H$ as described in section \ref{sec: method}) for a sample of instances showed that the model attends to only the correct idiom (in relatively shorter sentences) or to many phrases in a sentence (longer or those with literal phrases). In the sentence \textit{but they'd had \underline{a thorough look through his life} and just to be sure and \underline{hit the jackpot} entirely \underline{by chance}}, underlined phrases are those with high attention and \textit{hit the jackpot} was correctly selected as the output\footnote{Owing to space constraints we were unable to present detailed illustrations of attention matrices to make our point.}. In some instances of incorrect prediction, i.e., incompletely identified  IE tokens  or wrongly predicting the sentence to be literal, we found that the model still attended to the correct phrase. 
{We hypothesize that the two attention flow layers have a hierarchical relation in their functions: the first attention flow layer, using static word embeddings and their POS tags, identifies candidate phrases that could have idiomatic meanings; and the second attention flow layer, by checking for SC, identifies the idiomatic expression's span if it exists.}
Hence, accurate span prediction requires the model to (1) attend to the right tokens, (2) generate/extract meaningful token representations (from the attention phase), and then (3) correctly classify the tokens. Based on the fact that the model is attending to the phrases correctly, future studies should improve upon  the prediction phase using models that more efficiently leverage the features for improved token classification. 

\begin{table*}[!th]
\centering
\caption{Case studies on the \our's idiom identification. The ground truth PIEs are in italic and colored green in sentences. The Error Type column lists the name of the error and their percentage (Pct.) in parenthesis. The percentage is obtained by manually categorizing 300 incorrect samples.}
\label{tab: case_study}
\resizebox{\textwidth}{!}{
\small
\begin{tabular}{|c|p{27mm}|p{80mm}|p{30mm}|}
\hline
\multicolumn{1}{|l|}{\textbf{Case \#}} & \textbf{Error Type (Pct.)} & \textbf{Sentence with {\color[HTML]{009901}\textit{PIE}}} & \textbf{Prediction} \\ \hline
1 & Alternative (9.7\%) & But an {\color[HTML]{009901}\textit{on-the-ball}} whisky shop could make a killing with its special ec-label malt scotch at £27.70 a bottle. & make a killing \\ \hline
2 & Partial (29.3\%) & Dragons can lie for dark centuries brooding over their treasures , bedding down on frozen flames that will never {\color[HTML]{009901}\textit{see the light of day}}. & of day \\ \hline
3 & Meaningful (4.3\%) & Given a method, we can avoid mistaken ideas which, confirmed by the authority of the past, have {\color[HTML]{009901}\textit{taken deep root}}, like weeds in men's minds. & weeds in men's minds \\ \hline
4 & Literal (8.0\%) & If you must jump {\color[HTML]{009901}\textit{out of the loop}} , you should use until true to ``pop'' the stack. & out of the loop \\ \hline
5 & Missing (42.3\%) & We have {\color[HTML]{009901}\textit{friends in high places}}, they said. & Empty string \\ \hline
6 & Other (6.3\%) & With the chips down , we had to {\color[HTML]{009901}\textit{dig down}}. & With down \\ \hline
\end{tabular}%
}
\end{table*}

Moreover, we present case studies on the wrongly predicted instances from the MAGPIE type-aware split. {Toward this, we randomly sample ~25\%  of incorrectly predicted instances by \our (300 instances), and  group them into 6 case types: (1) \textit{alternative}, (2) \textit{partial}, (3) \textit{meaningful}, (4) \textit{literal}, (5) \textit{missing}, and (6) \textit{other}.} These are shown in Table~\ref{tab: case_study} and we discuss them below. 

{
Case 1 is the ``alternative'' case, which is a common `mis-identification' where \our  only identifies one of the IEs when multiple IEs are present; hence, the model detects the alternative IE to the IE originally labeled as the ground truth. Strictly speaking,  this is not a limitation of our method but rather an artifact of the  available dataset; all the datasets used in our experiments only label at most one PIE for each sentence even when there may be more than one. 
Case 2 is the ``partial'' case, which is another common wrong prediction where only a portion of the idiom span is recognized, i.e., the boundary of the entire idiom is not precisely localized.
Case 3 is the ``meaningful'' case in which  \our  identifies figurative expressions instead of the ground truth idiom (and in this sense relates to Case 1 above). As an example, when the ground truth is \textit{taken deep root}, \our  identifies \textit{weeds in men's minds,} which is clearly used metaphorically and so could have been an acceptable answer. Since the idioms are unknown to \our  during test time, we argue that, as in Case 1, the identification is still meaningful, although the detected phrases are not exactly the same  as the ground truth.
Case 4 is the ``literal'' case in which \our identifies a PIE that is actually used in the literal sense. 
Case 5, the ``missing'' case, is the opposite of Case 4,  where  \our  fails to recognize the presence of an idiom completely and returns only an empty string.
Case 6 is the final error type ``other'' in which \our returns  words or phrases that are not meaningful or figurative, nor part of any PIEs. 
}

{After categorizing the 300 incorrect instances according to the above definitions, we found that 42.3\% of them are of the ``missing'' case and around 43.4\%  are samples with partially correct predictions or meaningful alternative predictions. Their detailed  breakdown is listed in Table~\ref{tab: case_study}. Tackling the erroneous cases will be a fruitful future endeavor.}

\section{Conclusion and Future Work} \label{sec: conclusion}
In this work, we studied how a neural architecture that fuses multiple levels of syntactic and semantic information of words can effectively perform idiomatic expression identification. Compared to competitive baselines, the proposed model yielded state-of-the-art performance on  PIEs that varied with respect to syntactic patterns, degree of compositionality and syntactic flexibility. A salient feature of the model is its ability to generalize to PIEs unseen in the training data.

Although the exploration in this work is limited to IEs, we made no idiom-specific assumptions in the model. Future directions should extend the study to nested and syntactically flexible PIEs (verbal MWEs) and other figurative/literal constructions such as metaphors---categories that were not sufficiently represented in the datasets considered in this study.  
Other concrete research directions include performing the task in cross- and multi-lingual settings.

\section*{Acknowledgements}
We thank the anonymous reviewers for their comments 
on earlier drafts that significantly helped improve this manuscript. This work was  supported by the IBM-ILLINOIS Center for Cognitive Computing Systems Research (C3SR)---a research collaboration as part of the IBM AI Horizons Network.

\bibliography{draft_paper}
\bibliographystyle{acl_natbib}

\end{document}